\documentclass{article}

\usepackage{microtype}
\usepackage{graphicx}
\usepackage{subcaption}
\usepackage{booktabs} 

\usepackage{hyperref}

\usepackage[preprint]{icml2026}

\usepackage{amsmath}
\usepackage{amssymb}
\usepackage{mathtools}
\usepackage{amsthm}

\usepackage[capitalize,noabbrev]{cleveref}

\theoremstyle{plain}

\theoremstyle{definition}

\theoremstyle{remark}

\usepackage[textsize=tiny]{todonotes}

\icmltitlerunning{Submission and Formatting Instructions for ICML 2026}

\begin{document}

\twocolumn[
  \icmltitle{VLM-Guided Iterative Refinement for Surgical Image Segmentation with Foundation Models}



  \icmlsetsymbol{equal}{*}

  \begin{icmlauthorlist}
    \icmlauthor{Ange Lou}{equal,comp}
    \icmlauthor{Yamin Li}{equal,comp}
    \icmlauthor{Qi Chang}{psu}
    \icmlauthor{Nan Xi}{vcu}
    \icmlauthor{Luyuan Xie}{pku}
    \icmlauthor{Zichao Li}{ucsd}
    \icmlauthor{Tianyu Luan}{yyy}
  \end{icmlauthorlist}

  \icmlaffiliation{yyy}{School of Automation Science and Engineering, South China University of Technology, Guangzhou, China}
  \icmlaffiliation{comp}{Vanderbilt University, Nashville, TN, USA}
  \icmlaffiliation{psu}{The Pennsylvania State University, University Park, PA , USA}
  \icmlaffiliation{pku}{Peking University, Beijing , China}
  \icmlaffiliation{vcu}{Virginia Commonwealth University, Richmond, VA, USA}
  \icmlaffiliation{ucsd}{University of California San Diego, San Diego, CA, USA}

  \icmlcorrespondingauthor{Tianyu Luan}{luanty.henry@gmail.com}

  \icmlkeywords{Machine Learning, ICML}

  \vskip 0.3in
]



\printAffiliationsAndNotice{}  

\begin{abstract}
Surgical image segmentation is essential for robot-assisted surgery and intraoperative guidance. However, existing methods are constrained to predefined categories, produce one-shot predictions without adaptive refinement, and lack mechanisms for clinician interaction. We propose IR-SIS, an iterative refinement system for surgical image segmentation that accepts natural language descriptions. IR-SIS leverages a fine-tuned SAM3 for initial segmentation, employs a Vision-Language Model to detect instruments and assess segmentation quality, and applies an agentic workflow that adaptively selects refinement strategies. The system supports clinician-in-the-loop interaction through natural language feedback. We also construct a multi-granularity language-annotated dataset from EndoVis2017 and EndoVis2018 benchmarks. Experiments demonstrate state-of-the-art performance on both in-domain and out-of-distribution data, with clinician interaction providing additional improvements. Our work establishes the first language-based surgical segmentation framework with adaptive self-refinement capabilities.
\end{abstract}

\begin{figure}[ht]
    \centering
    \includegraphics[width=1\linewidth]{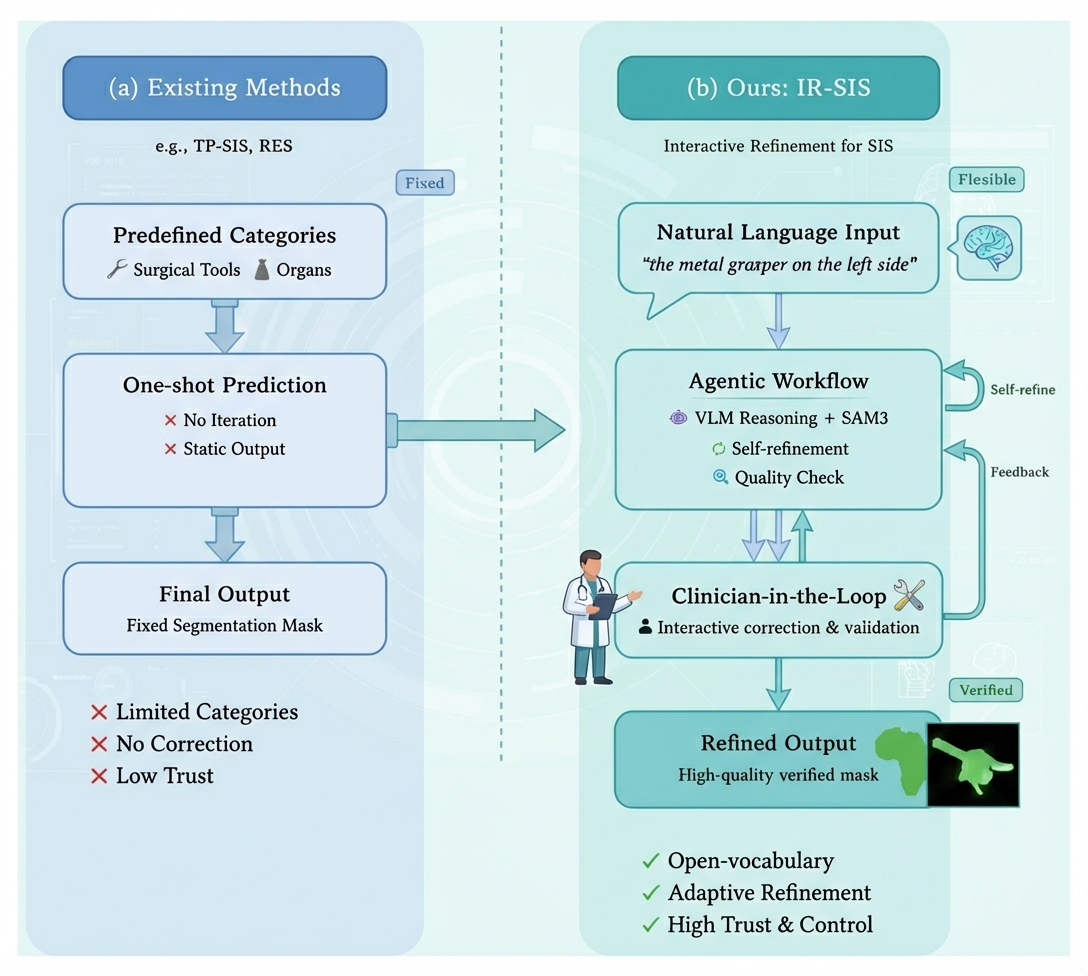}
    \caption{Existing language-based surgical segmentation methods face two key limitations: a) They are restricted to predefined instrument categories (e.g., TP-SIS), producing one-shot predictions without adaptive refinement or clinician interaction. (b) Our approach IR-SIS enables flexible natural language descriptions, employs an agentic workflow for adaptive self-refinement, and supports clinician-in-the-loop interaction for quality improvement.}
    \label{fig:pipeline}
\end{figure}

\section{Introduction}
\label{sec:intro}

Surgical image segmentation is a fundamental capability for computer-assisted intervention systems. Accurate segmentation of surgical instruments and tissues enables surgical navigation, robotic automation, and post-operative analysis. However, real surgical environments present substantial challenges: new instruments are continuously introduced, different procedures require different segmentation targets, and clinical workflows demand both precision and adaptability. These requirements call for segmentation systems that can handle diverse, evolving instrument types through flexible user interaction.

Existing surgical image segmentation methods fall short of these requirements in two fundamental ways. Traditional approaches train on predefined instrument categories, causing performance degradation when encountering instruments absent from training data. And this is a frequent occurrence given the rapid evolution of surgical tools~\cite{zhou2023text}. Recent language-based methods such as TP-SIS~\cite{zhou2023text} represent progress by accepting text prompts instead of category IDs. However, TP-SIS remains constrained to a fixed vocabulary of instrument types (e.g., ``forceps'', ``scissors'') and cannot interpret free-form descriptions such as ``the instrument on the left'' or ``the grasping tool near the tissue.'' More critically, both traditional and language-based methods perform one-shot prediction: the model produces a segmentation mask in a single forward pass with no mechanism for quality assessment, iterative refinement, or incorporation of clinician feedback. When initial segmentation contains errors, whether from ambiguous instrument boundaries, occlusion, or unusual lighting, these methods offer no path to correction. The absence of adaptive refinement and human-in-the-loop interaction limits their reliability in clinical settings where segmentation errors can have downstream consequences.

We observe that human experts do not annotate surgical images in a single pass. Instead, they follow an iterative process: initial annotation, quality assessment, targeted correction, and confirmation. This observation motivates our core insight: segmentation systems should emulate this iterative workflow rather than rely on one-shot prediction. Vision-Language Models (VLMs) now possess strong visual understanding and reasoning capabilities that extend beyond simple classification. A VLM can detect instruments in an image, localize them with bounding boxes, and assess whether a given segmentation mask adequately covers the detected objects. By treating the VLM as a quality evaluator, we can construct an agentic workflow that iteratively refines segmentation based on quantitative quality metrics. This architecture also provides a natural interface for clinician-in-the-loop interaction: when the VLM's automatic assessment is insufficient, clinicians can provide targeted feedback that guides subsequent refinement iterations.

Based on this insight, we propose IR-SIS (Iterative Refinement for Surgical Image Segmentation), a VLM-guided framework that transforms surgical image segmentation from one-shot prediction to adaptive iterative refinement. IR-SIS comprises four components: (1) a SAM-based segmenter~\cite{kirillov2023segment,ravi2024sam2} fine-tuned on our multi-level language annotation dataset, which performs initial language-based segmentation; (2) a VLM instrument detector that identifies all surgical instruments and outputs bounding boxes; (3) a quality evaluator that computes mask coverage and mask-box overlap metrics; and (4) an agentic refinement workflow that selects between trust-initial and multi-instrument strategies based on quality scores. When quality metrics fall below threshold, the system uses detected bounding boxes as prompts to re-segment individual instruments and iteratively improves the result. Clinicians can intervene at any iteration to provide corrective feedback.

Our contributions are as follows:
\begin{itemize}
    \item \textbf{Language-based surgical segmentation system.} We propose a segmentation system that accepts natural language descriptions of target instruments, enabling clinicians to specify segmentation targets flexibly rather than selecting from predefined categories. This accommodates diverse clinical vocabulary and novel instrument types without retraining.
    \item \textbf{Agentic iterative refinement workflow.} We design a VLM-guided workflow that assesses segmentation quality and performs adaptive refinement, upgrading the paradigm from one-shot prediction to intelligent iteration. The workflow supports clinician-in-the-loop interaction, allowing human feedback to guide refinement when automatic assessment is insufficient.
    \item \textbf{Multi-level language annotation dataset.} We construct a dataset with three-level granularity annotations (general, category, specific) covering EndoVis2017 and EndoVis2018 benchmarks. This enables training models that understand queries at different specificity levels, from ``surgical instrument'' to ``bipolar forceps.''
    \item \textbf{Comprehensive experimental validation.} We evaluate IR-SIS on in-domain and out-of-distribution data, demonstrating that the agentic workflow improves segmentation quality, clinician interaction provides additional gains, and models trained on our dataset exhibit stronger generalization to unseen instruments.
\end{itemize}

\begin{figure*}[ht]
    \centering
    \includegraphics[width=1\linewidth]{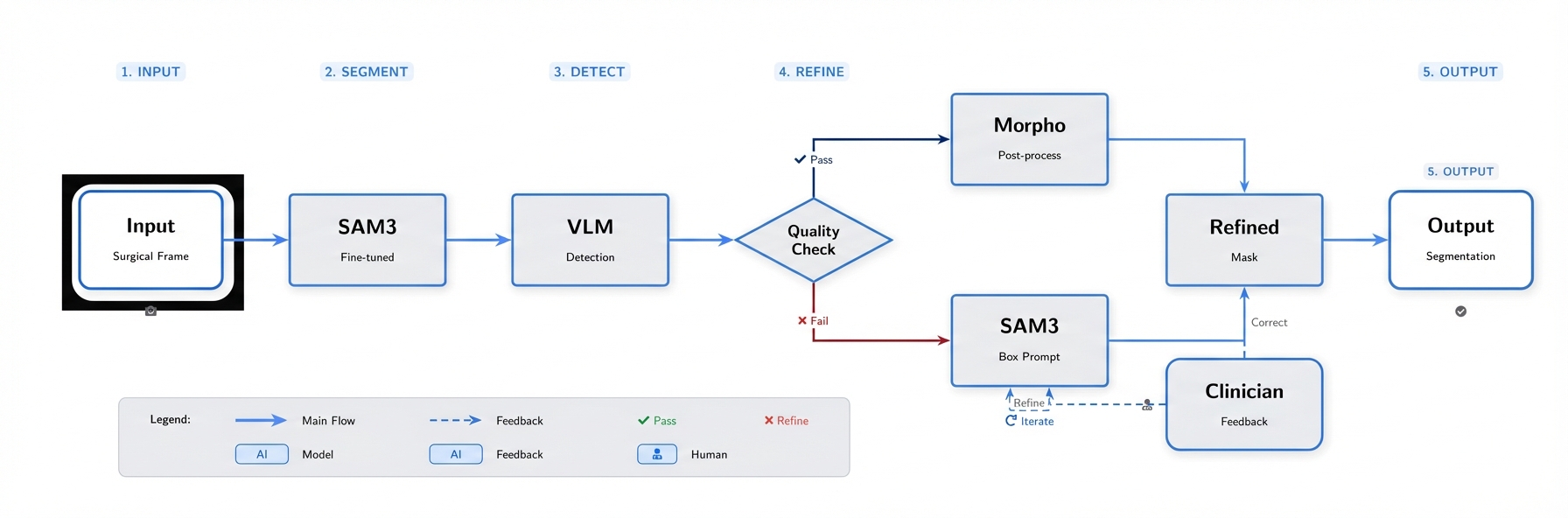}
    \caption{IR-SIS Pipeline Overview. We first fine-tune SAM3 on our dataset to perform initial segmentation on input surgical images. Subsequently, a VLM is employed to detect all surgical instruments. Segmentation quality is then evaluated based on mask coverage and mask-box overlap ratio: if the quality meets the criteria, only morphological post-processing is applied; otherwise, each instrument is re-segmented using box prompts followed by iterative refinement. Additionally, clinicians can provide feedback during the iterative process to guide the refinement direction, ultimately yielding refined segmentation results. }
    \label{fig:pipeline}
\end{figure*}

\section{Related Work}
\label{sec:related}
\subsection{Surgical Instrument Segmentation}
\label{subsec:surgical-seg}

Deep learning has become the dominant approach for surgical instrument segmentation. Early methods built upon U-Net and its variants achieved strong performance on the EndoVis benchmark datasets~\cite{shvets2018ternausnet,allan2019endovis2017}. ISINet~\cite{gonzalez2020isinet} introduced instance-level segmentation with temporal consistency modeling, while TraSeTR~\cite{zhao2022trasetr} leveraged tracking cues within a transformer framework to improve instance-level predictions.
The Segment Anything Model (SAM)~\cite{kirillov2023segment} demonstrated powerful zero-shot segmentation capabilities, prompting adaptations for medical imaging. MedSAM~\cite{ma2024medsam} fine-tuned SAM on large-scale medical data covering diverse modalities. SurgicalSAM~\cite{yue2024surgicalsam} and AdaptiveSAM~\cite{paranjape2024adaptivesam} specifically targeted surgical scene segmentation with efficient tuning strategies. However, these methods require explicit geometric prompts (points, boxes) and lack the ability to understand natural language descriptions of target instruments.

\subsection{Language-Guided Image Segmentation}
\label{subsec:lang-seg}

Contrastive Language-Image Pretraining (CLIP)~\cite{radford2021clip} enabled vision-language models to understand semantic relationships between images and text. CRIS~\cite{wang2022cris} and CLIPSeg~\cite{luddecke2022clipseg} transferred CLIP's semantic knowledge to pixel-level segmentation, while LAVT~\cite{yang2022lavt} integrated linguistic information directly into vision transformer features.
In the surgical domain, TP-SIS~\cite{zhou2023text} pioneered text-promptable instrument segmentation using a mixture of prompts mechanism. However, TP-SIS and related methods remain constrained to fixed category prompts and produce one-shot predictions without iterative refinement. Interactive segmentation approaches~\cite{marinov2024interactive,zhu2024meduhip} incorporate user feedback for iterative improvement, but require manual intervention at each iteration.
Our IR-SIS differs from prior work in three aspects: (1) it accepts flexible natural language descriptions rather than fixed category prompts, (2) it employs a VLM-guided agentic workflow for autonomous quality assessment and adaptive refinement, and (3) it supports optional clinician-in-the-loop interaction for guided improvement when needed.

\section{Method}
\label{sec:method}

\subsection{Problem Formulation}
\label{subsec:problem}

We address the task of language-based surgical image segmentation. Given a surgical image $I \in \mathbb{R}^{H \times W \times 3}$ and a natural language query $q$ describing the target instrument, the goal is to predict a binary segmentation mask $M \in \{0,1\}^{H \times W}$ that identifies all pixels belonging to the queried object. Unlike traditional surgical segmentation methods that operate on fixed instrument categories, our formulation accepts free-form text queries at multiple granularity levels from general descriptions (``surgical instrument'') to specific identifiers (``bipolar forceps''). We formalize the objective as:
\begin{equation}
\hat{M} = \arg\max_M p(M | I, q; \theta)
\label{eq:objective}
\end{equation}
where $\theta$ represents the model parameters. To achieve this goal while enabling adaptive refinement and clinician interaction, we propose IR-SIS, a VLM-guided iterative refinement system.

\subsection{Method Overview}
\label{subsec:overview}

As illustrated in \cref{fig:pipeline}, IR-SIS consists of four stages that transform surgical image segmentation from one-shot prediction to adaptive iterative refinement. First, our language-guided segmenter, built on a fine-tuned SAM architecture~\cite{kirillov2023segment,ravi2024sam2}, generates an initial segmentation mask $M_0$ from the image and text query. Second, a Vision-Language Model (VLM) detector~\cite{bai2023qwenvl} identifies all visible surgical instruments in the image and outputs their bounding boxes $B = \{b_1, \ldots, b_n\}$. Third, a quality evaluator computes mask coverage and mask-box overlap metrics to assess whether the initial segmentation adequately captures the target instruments. Fourth, an agentic refinement workflow uses these quality scores to select between strategies: if quality is sufficient, the system applies morphological post-processing and outputs the result; otherwise, it iteratively re-segments low-quality regions using box prompts until the quality threshold is met or maximum iterations are reached. Clinicians can provide feedback at any iteration to guide the refinement direction.
We now describe each component in detail.

\subsection{IR-SIS Pipeline}
\label{subsec:pipeline}

\textbf{Language-Guided Segmenter.} Existing surgical segmentation methods such as TP-SIS~\cite{zhou2023text} support only fixed-vocabulary prompts and cannot generalize to free-form language descriptions or novel instrument types. To address this limitation, we fine-tune a SAM-based architecture~\cite{kirillov2023segment,ravi2024sam2} using a hierarchical learning rate strategy that preserves pre-trained features while adapting to surgical image segmentation.

The vision backbone uses a low learning rate with layer decay to retain pre-trained visual representations. The text encoder remains frozen to leverage its pre-trained language understanding capabilities. The transformer decoder receives a higher learning rate to learn the segmentation task. This configuration enables the model to adapt to surgical domains without catastrophically forgetting general visual knowledge.

To enhance generalization across query granularities, we train with multi-level prompts. Each annotated instrument generates three training samples at different specificity levels: Level 0 for general queries (e.g., ``surgical instrument''), Level 1 for category-level queries (e.g., ``forceps''), and Level 2 for specific instrument queries (e.g., ``bipolar forceps''). The segmenter produces an initial mask:
\begin{equation}
M_0 = f_{\text{SAM}}(I, q; \theta)
\label{eq:initial_mask}
\end{equation}
where $\theta$ represents the fine-tuned parameters.

\textbf{VLM Instrument Detector.} The initial segmentation may miss instruments or include multiple objects beyond the query target. To provide a global view of all instruments present in the image, we employ a VLM~\cite{bai2023qwenvl} to detect visible surgical instruments. Given the image $I$ and a detection prompt $p_{\text{detect}}$, the VLM outputs bounding boxes for all detected instruments:
\begin{equation}
B = \{b_i\}_{i=1}^n = \text{VLM}(I, p_{\text{detect}})
\label{eq:detection}
\end{equation}
where $n$ is the number of detected instruments. These bounding boxes serve two purposes: they enable quality assessment by providing reference regions, and they provide prompts for iterative refinement when needed.

\textbf{Quality Evaluator.} Given the detection results, we quantify the quality of the initial segmentation to determine whether refinement is necessary. We define two complementary metrics. \emph{Mask Coverage} measures what fraction of detected instrument regions are covered by the segmentation:
\begin{equation}
C = \frac{|M \cap \bigcup_{i} b_i|}{|\bigcup_{i} b_i|}
\label{eq:coverage}
\end{equation}
\emph{Mask-Box Overlap} measures how well the segmentation aligns with the target bounding box:
\begin{equation}
O = \frac{|M \cap b_{\text{target}}|}{|b_{\text{target}}|}
\label{eq:overlap}
\end{equation}
The quality indicator combines these metrics:
\begin{equation}
\mathcal{S} = H(C > \tau_c) \land H(O > \tau_o)
\label{eq:quality}
\end{equation}
where $\tau_c$ and $\tau_o$ are thresholds tuned on a validation set. $H(\cdot)$ is a step function that outputs 1 when the value is above the threshold and 0 otherwise, acting as a simple on–off switch for quality.

\begin{figure}[t]
    \centering
    \includegraphics[width=0.99\linewidth]{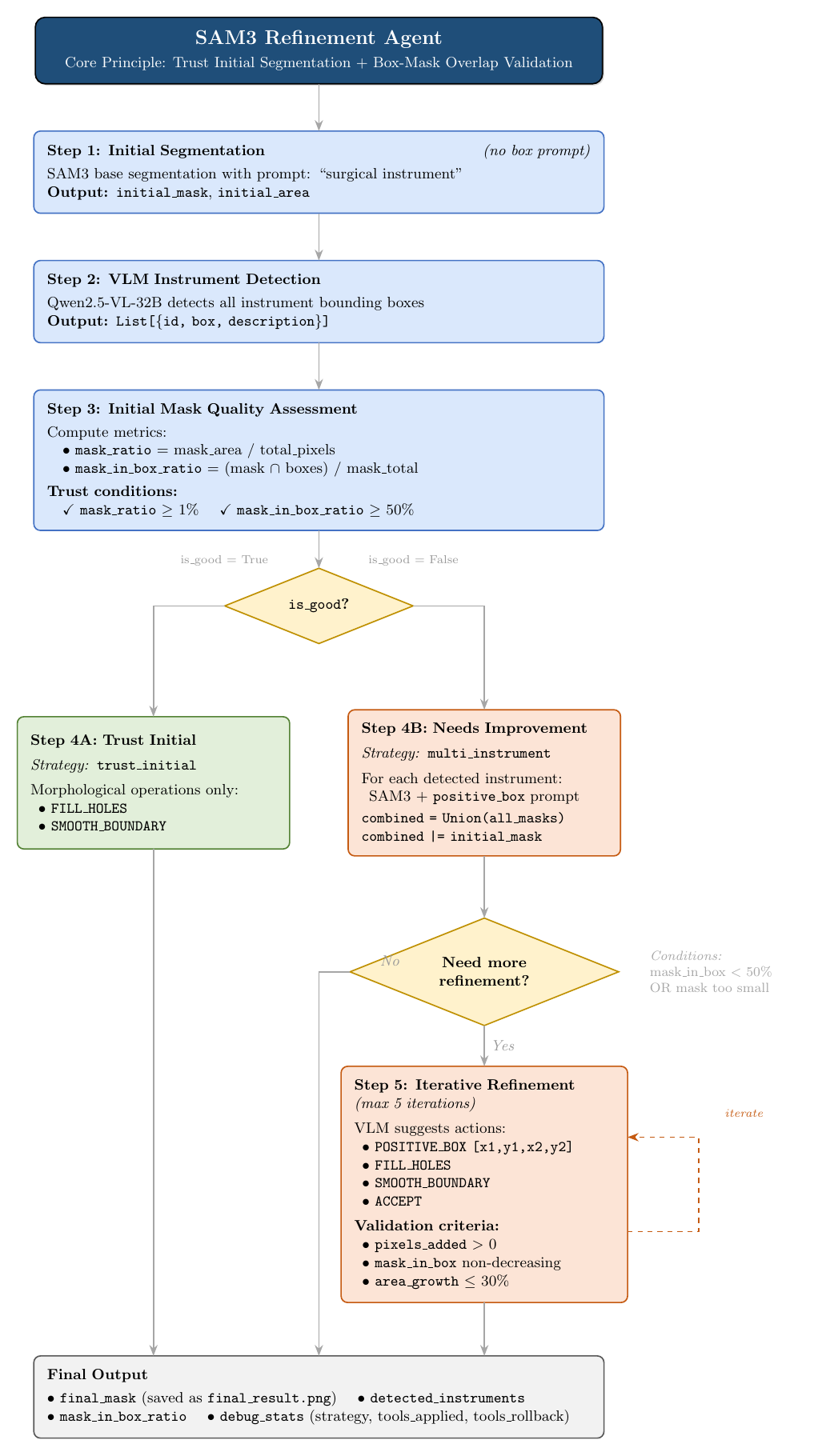}
    \caption{Detailed workflow of the SAM3 Refinement Agent. The agent adaptively selects between trust-initial and multi-instrument strategies based on mask-box overlap quality.}
    \label{fig:detail}
\end{figure}

\begin{table*}[ht]
\caption{Comparison between our method and other state-of-the-art methods on the EndoVis2018 dataset.}
\label{tab:endovis2018}
\centering
\resizebox{\linewidth}{!}{
\begin{tabular}{l|cc|cccccccc|c}
\toprule
Method & Ch\_IoU & BF & PF & LND & SI & CA & MCS & UP & mc\_IoU \\
\midrule
TernausNet-11 \cite{iglovikov2018ternausnet} & 46.22  & 44.20 & 4.67 & 0.00 & 0.00 & 0.00 & 50.44 & 0.00 & 14.19 \\
MF-TAPNet \cite{jin2019incorporating} & 67.87  & 69.23 & 6.10 & 11.68 & 14.00 & 0.91 & 70.24 & 0.57 & 24.68 \\
ISINet \cite{gonzalez2020isinet} & 73.03  & 73.83 & 48.61 & 30.98 & 37.68 & 0.00 & 88.16 & 2.16 & 40.21 \\
TraSeTR \cite{zhao2022trasetr} & 76.20  & 76.30 & 53.30 & 46.50 & 40.60 & 13.90 & 86.30 & 17.50 & 47.77 \\
S3Net \cite{xue2022s3net} & 75.81  & 77.22 & 50.87 & 19.83 & 50.59 & 0.00 & 92.12 & 7.44 & 42.58 \\
MATIS \cite{ayobi2023matis} & 84.26  & 83.52 & 41.90 & 66.18 & 70.57 & 0.00 & \textbf{92.96} & 23.13 & 54.04 \\
\midrule
CRIS \cite{wang2022cris} & 74.10  & 73.58 & 58.20 & 47.64 & 72.14 & 4.56 & 45.99 & 20.18 & 46.04 \\
CLIPSeg \cite{luddecke2022clipseg} & 74.95  & 67.25 & 39.59 & 36.72 & 47.27 & 2.92 & 79.96 & 4.22 & 39.7 \\
TP-SIS(448) \cite{zhou2023text} & 82.67  & 81.53 & 70.18 & 71.54 & 90.58 & 21.46 & 65.57 & 57.51 & 65.48 \\
TP-SIS(896) \cite{zhou2023text} & 84.92 & 84.28 & 73.18 & 78.88 & 92.20 & 23.73 & 66.67 & 39.12 & 65.44 \\
\textbf{Ours} & \textbf{85.44} & \textbf{87.99} & \textbf{82.69} & \textbf{49.65} & \textbf{86.41} & \textbf{31.12} & 92.86 & \textbf{79.39} & \textbf{72.87} \\
\bottomrule
\end{tabular}
}
\end{table*}

\begin{table}[ht]
\centering
\caption{VLM Model Comparison on Kvasir-Instrument Dataset}
\resizebox{\linewidth}{!}{
\begin{tabular}{lcc}
\toprule
VLM Model & mean IoU & Improvement \\
\midrule
Baseline (SAM3 finetuned) &71.08 & N/A \\
Qwen2.5-VL-32B-Instruct &  77.95 & +6.87 \\
Qwen3-VL-32B-Instruct (Ours) &  78.90 & +7.82 \\
\bottomrule
\end{tabular}
}
\label{tab:ood}
\end{table}

\textbf{Agentic Refinement Workflow.} One-shot prediction struggles with complex surgical scenes involving occlusion, multiple instruments, or low contrast. Based on the quality assessment, our agentic workflow adaptively selects the appropriate refinement strategy. The psudeo-code-level process of our workflow is illustrated in \cref{fig:detail}.

When quality metrics indicate the initial segmentation is adequate ($\mathcal{S} = 1$), the system applies morphological post-processing only. Opening operations remove small noise artifacts, and closing operations fill holes: $M_{\text{final}} = \text{Morph}(M_0)$.
When quality is insufficient ($\mathcal{S} = 0$), the system identifies which bounding boxes have low-quality segmentation and re-segments those regions using box prompts. The process iterates until quality thresholds are met or maximum iterations are reached:
\begin{equation}
M^{(t+1)} =
\begin{cases}
\text{Morph}(M^{(t)}) & \text{if } \mathcal{S}^{(t)} = 1 \\
\text{SAM}(I, b_{\text{low}}) \cup M^{(t)}_{\text{good}} & \text{if } \mathcal{S}^{(t)} = 0
\end{cases}
\label{eq:refinement}
\end{equation}
where $b_{\text{low}}$ denotes the bounding box with low-quality segmentation and $M^{(t)}_{\text{good}}$ is the high-quality portion preserved from iteration $t$.

The workflow supports clinician intervention at any iteration through three interaction modes: (1) box prompts to specify regions of interest or missed instruments, (2) language descriptions to supplement or correct the query, and (3) reference annotations to guide the segmentation. Clinician feedback takes priority over automatic VLM assessment, ensuring clinical controllability.

\subsection{Training}
\label{subsec:training}

\textbf{Loss Function.} We train the language-guided segmenter with a composite loss function:
\begin{equation}
\mathcal{L}_{\text{total}} = \lambda_{\text{mask}} \cdot \mathcal{L}_{\text{mask}} + \lambda_{\text{dice}} \cdot \mathcal{L}_{\text{dice}} + \lambda_{\text{ce}} \cdot \mathcal{L}_{\text{ce}} + \lambda_{\text{presence}} \cdot \mathcal{L}_{\text{presence}}
\label{eq:loss}
\end{equation}
The mask loss $\mathcal{L}_{\text{mask}}$ is a focal loss ($\alpha$=0.25, $\gamma$=2) that handles foreground/background imbalance. The dice loss $\mathcal{L}_{\text{dice}}$ improves small object segmentation. The cross-entropy loss $\mathcal{L}_{\text{ce}}$ supervises query-object matching classification. The presence loss $\mathcal{L}_{\text{presence}}$ predicts whether the queried object exists in the image. In our experiments, we set $\lambda_{\text{mask}}$, $\lambda_{\text{dice}}$, $\lambda_{\text{ce}}$, and $\lambda_{\text{presence}}$ to 5, 1, 2, 2, respectively.

\textbf{Hierarchical Learning Rate Strategy.} Different model components receive different learning rates to balance adaptation and preservation of pre-trained knowledge. The transformer decoder uses a learning rate of $8 \times 10^{-5}$ for task-specific learning. The vision backbone uses $2.5 \times 10^{-5}$ with layer decay (0.98) to preserve pre-trained features. The text encoder remains frozen to retain language understanding capabilities.

\textbf{Matching Strategy.} We employ Hungarian matching combined with one-to-many matching (top-k=4, weight=2.0) to balance positive and negative samples and improve training stability.

\textbf{Data Augmentation.} We apply domain-specific augmentations to improve robustness to surgical imaging conditions: color jitter for color variability, gamma correction ($\gamma \in [0.8, 1.2]$) for exposure adaptation, and specular highlight and random shadow augmentations to simulate surgical scene lighting conditions.

\textbf{Quality Evaluation Thresholds.} The coverage threshold $\tau_c$ and overlap threshold $\tau_o$ are tuned on a validation set to balance precision and recall in the quality assessment. Morphological operations use kernel sizes selected based on typical instrument dimensions in the training data.


\begin{figure*}[t]
    \centering
    \includegraphics[width=0.85\linewidth]{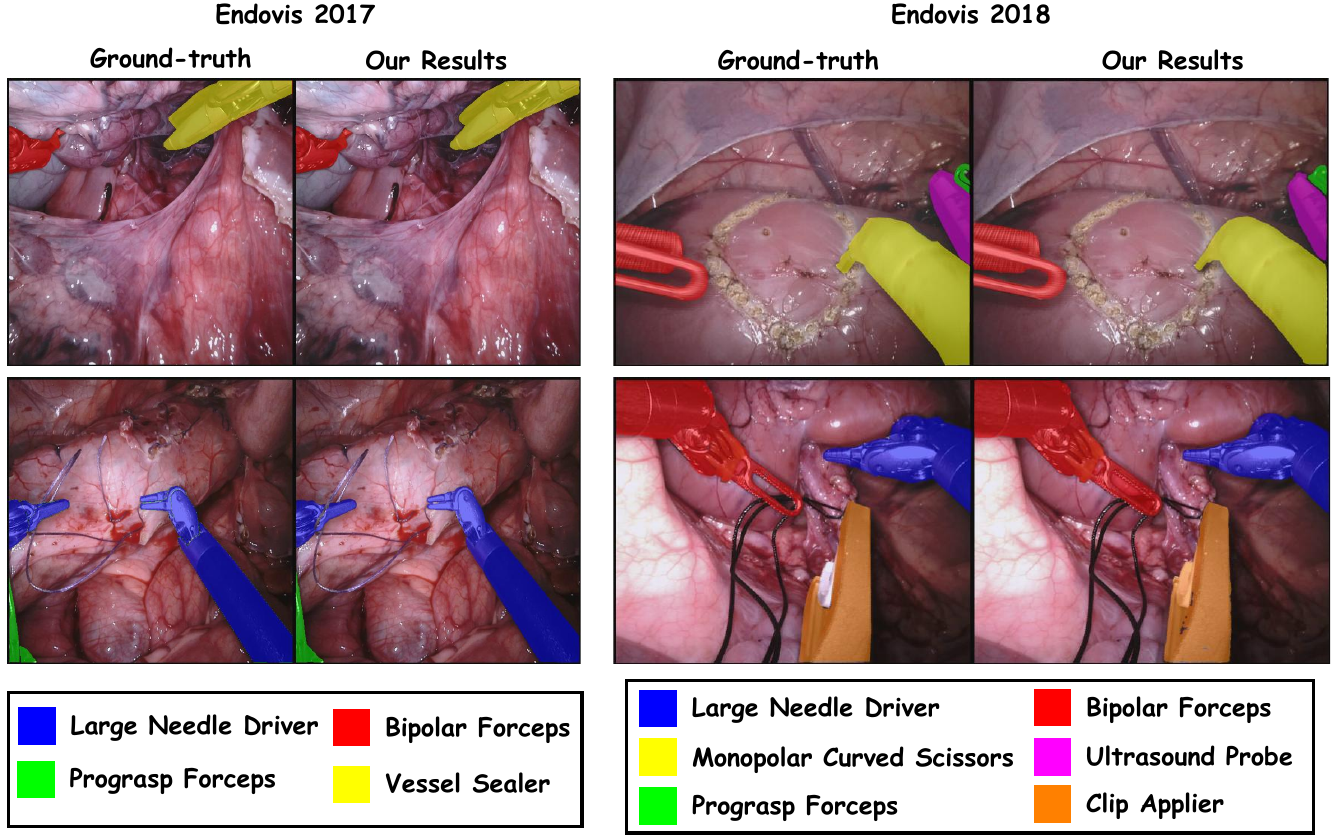}
    \caption{Text prompt segmentation results on Endovis2017 and Endovis2018 Datasets}
    \label{fig:in}
\end{figure*}

\begin{figure*}[t]
    \centering
    \includegraphics[width=1\linewidth]{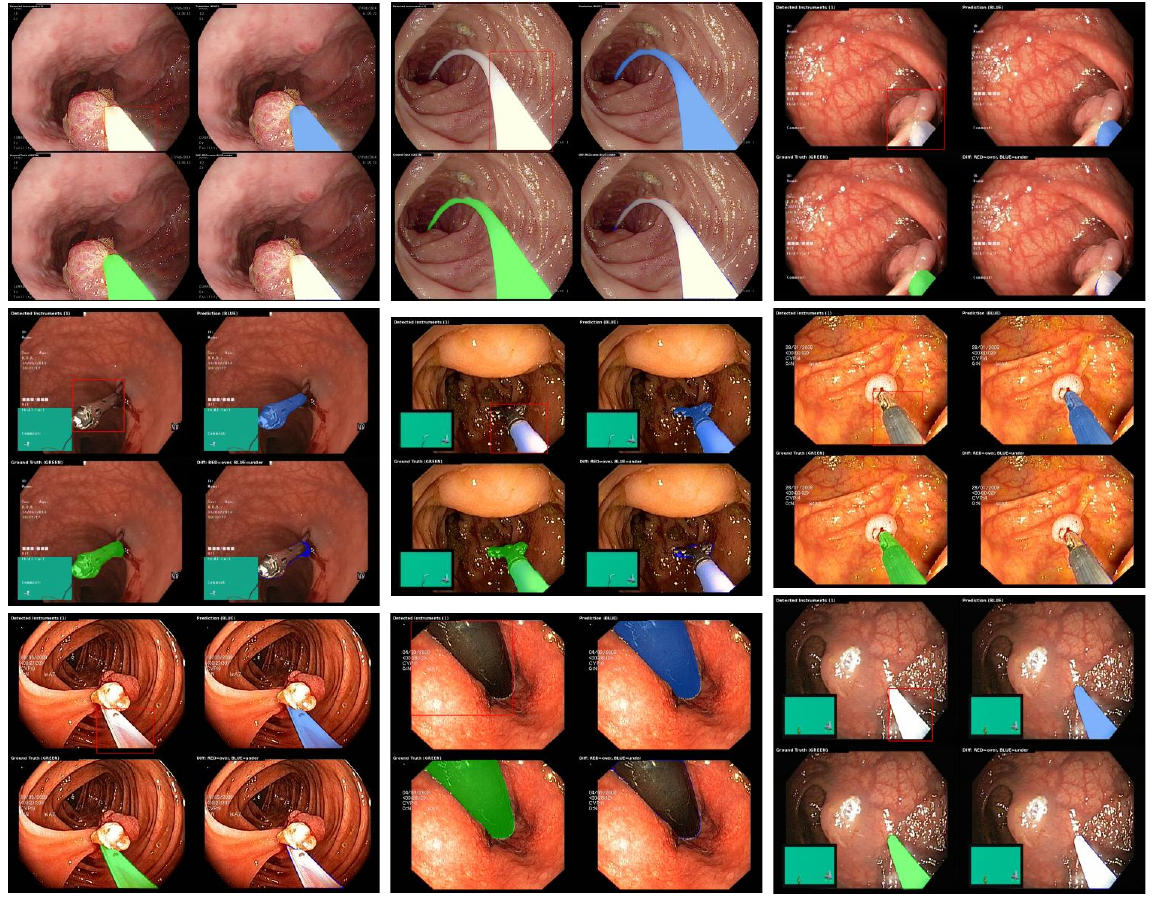}
    \caption{Text prompt segmentation results on Kvasir-instrument dataset}
    \label{fig:ood}
\end{figure*}

\section{Experiments}
\label{sec:experiments}

We evaluate IR-SIS on both in-domain and out-of-distribution surgical image segmentation tasks. Our experiments demonstrate the effectiveness of the proposed agentic workflow and the benefits of clinician-in-the-loop interaction.

\subsection{Datasets}
\label{subsec:datasets}

\textbf{Training Data.} We train our model on a combined dataset from EndoVis2017 and EndoVis2018 challenges. EndoVis2017 contains 1,778 images from 8 robotic surgery sequences with multiple instrument types. EndoVis2018 provides 1,630 images from 15 surgical videos with diverse instrument appearances. Both datasets are annotated with our multi-level language annotations (L0-L2), resulting in 3,067 training images with 19,494 annotations after the 3x expansion from multi-level prompts.

\textbf{Evaluation Data.} For in-domain evaluation, we use the test sets from EndoVis2017 and EndoVis2018, comprising 341 images. For out-of-distribution evaluation, we use Kvasir-Instrument, which contains gastrointestinal endoscopy images with different instrument types and visual characteristics not seen during training.

\subsection{Evaluation Metrics}
\label{subsec:metrics}

We adopt two standard segmentation metrics. \textbf{Dice Score (DSC)} measures the overlap between predicted and ground truth masks, computed as $\text{DSC} = 2|A \cap B| / (|A| + |B|)$. \textbf{Intersection over Union (IoU)} provides a complementary measure of segmentation quality, defined as $\text{IoU} = |A \cap B| / |A \cup B|$. Both metrics range from 0 to 1, with higher values indicating better segmentation performance.

\subsection{Implementation Details}
\label{subsec:implementation}

\textbf{Model Configuration.} We fine-tune SAM3 as our base segmentation model with an input resolution of $1008 \times 1008$. For the VLM component, we use Qwen-2.5-VL-32B-instruct as the default instrument detector, with comparisons to GPT-4o and other models.

\textbf{Training Settings.} We train on 4 NVIDIA A6000 GPUs (48GB each) with a total batch size of 64 (16 per GPU) using FP16 mixed precision. We employ layerwise learning rates: $8 \times 10^{-5}$ for the Transformer Decoder, $2.5 \times 10^{-5}$ for the Vision Backbone with layer decay of 0.98, and freeze the Text Encoder. The total loss combines Mask Focal Loss (weight=50, $\alpha$=0.25, $\gamma$=2), Dice Loss (weight=10), Classification Focal Loss (weight=20), and Presence Loss (weight=20). Training follows a 90-epoch schedule: 30 epochs warmup, 30 epochs decay, and 30 epochs cooldown.

\textbf{Data Augmentation.} We apply color jitter ($p$=0.8), gamma correction ($p$=0.5), specular highlight simulation ($p$=0.3), and random shadow ($p$=0.3) to improve robustness to surgical imaging variations.

\textbf{Baselines.} We compare against: (1) SAM3 without fine-tuning, (2) SAM3 fine-tuned on our dataset without the agentic workflow, and (3) TP-SIS evaluated in zero-shot setting.

\subsection{Experimental Results}
\label{subsec:results}

\textbf{Main Comparison.} We evaluate IR-SIS against baselines on both in-domain and out-of-distribution data. On the in-domain EndoVis test sets (\cref{tab:endovis2018}), our method achieves substantial improvements over SAM3 origin and TP-SIS. The fine-tuning on our multi-level annotated dataset provides a strong foundation, and the agentic refinement workflow further improves segmentation quality by adaptively correcting initial predictions.

On the out-of-distribution Kvasir-Instrument dataset (\cref{tab:ood} and \cref{fig:ood}), IR-SIS demonstrates superior generalization compared to baselines. The VLM-guided iterative refinement is particularly effective in this setting, as it can detect and correct segmentation errors without requiring dataset-specific training. This validates our design choice of combining a fine-tuned segmentation model with an adaptive refinement mechanism.

\textbf{VLM Comparison.} We compare different VLMs as the instrument detector in our agentic workflow in \cref{tab:ood}. GPT-4o (API) achieves the highest detection accuracy but incurs API costs and latency. Qwen-2.5-VL-32B-instruct provides competitive performance with local deployment, offering a practical balance between accuracy and accessibility. The choice of VLM affects the final segmentation quality through its impact on instrument detection accuracy, which guides the iterative refinement process.

\begin{figure*}[htb]
    \centering
    \includegraphics[width=0.75\linewidth]{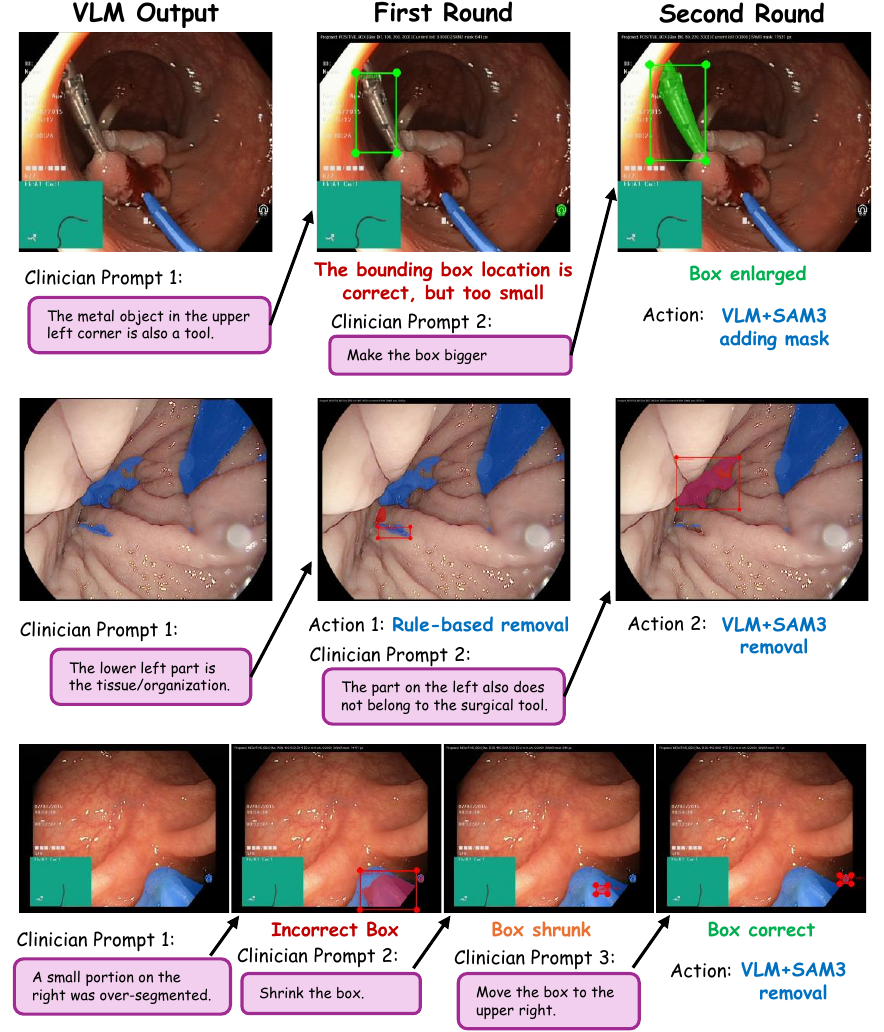}
    \caption{Clinician in the loop examples.}
    \label{fig:human}
\end{figure*}

\textbf{Language Diversity Demonstration.} IR-SIS supports flexible natural language queries at multiple granularity levels. In \cref{fig:human}, given the same surgical image, users can segment ``surgical instrument'' (all instruments), ``forceps'' (a specific category), ``bipolar forceps'' (a specific instrument), or even spatial descriptions like ``the instrument on the left side.'' This flexibility enables natural interaction for clinicians who may describe instruments differently based on their expertise and clinical context.

\textbf{Ablation Studies.} We conduct ablation experiments to validate key design choices. We evaluate in \cref{fig:human} the impact of clinician feedback on segmentation quality. When clinicians provide corrective prompts during the refinement process, the segmentation accuracy improves compared to VLM-only refinement. This demonstrates that our system can effectively leverage human expertise to handle challenging cases where automated refinement is insufficient.

\section{Conclusion}
\label{sec:conclusion}
We present IR-SIS, a VLM-guided framework that transforms surgical image segmentation from one-shot prediction to adaptive iterative refinement. By combining a language-based segmenter fine-tuned on multi-level annotations with an agentic workflow leveraging VLM-detected instruments and quality metrics, IR-SIS overcomes the limitations of predefined categories and lack of self-correction in existing methods. Experiments demonstrate improved segmentation quality through iterative refinement, additional gains from clinician-in-the-loop interaction, and strong generalization to unseen instruments.

\bibliography{example_paper}

@inproceedings{zhou2023text,
  title={Text Promptable Surgical Instrument Segmentation with Vision-Language Models},
  author={Zhou, Zijian and Alabi, Oluwatosin and Wei, Meng and Vercauteren, Tom and Shi, Miaojing},
  booktitle={Advances in Neural Information Processing Systems (NeurIPS)},
  volume={36},
  year={2023}
}

@inproceedings{kirillov2023segment,
  title={Segment Anything},
  author={Kirillov, Alexander and Mintun, Eric and Ravi, Nikhila and Mao, Hanzi and Rolland, Chloe and Gustafson, Laura and Xiao, Tete and Whitehead, Spencer and Berg, Alexander C. and Lo, Wan-Yen and Doll{\'a}r, Piotr and Girshick, Ross},
  booktitle={Proceedings of the IEEE/CVF International Conference on Computer Vision (ICCV)},
  pages={4015--4026},
  year={2023}
}

@article{ravi2024sam2,
  title={SAM 2: Segment Anything in Images and Videos},
  author={Ravi, Nikhila and Gabeur, Valentin and Hu, Yuan-Ting and Hu, Ronghang and Ryali, Chaitanya and Ma, Tengyu and Khedr, Haitham and R{\"a}dle, Roman and Rolland, Chloe and Gustafson, Laura and Mintun, Eric and Pan, Junting and Alwala, Kalyan Vasudev and Carion, Nicolas and Wu, Chao-Yuan and Girshick, Ross and Doll{\'a}r, Piotr and Feichtenhofer, Christoph},
  journal={arXiv preprint arXiv:2408.00714},
  year={2024}
}

@article{shvets2018ternausnet,
  title={Automatic Instrument Segmentation in Robot-Assisted Surgery Using Deep Learning},
  author={Shvets, Alexey A and Rakhlin, Alexander and Kalinin, Alexandr A and Iglovikov, Vladimir I},
  journal={arXiv preprint arXiv:1803.01207},
  year={2018}
}

@article{allan2019endovis2017,
  title={2017 Robotic Instrument Segmentation Challenge},
  author={Allan, Max and Shvets, Alex and Kurmann, Thomas and Zhang, Zichen and Duber, Rahul and Joos, Sebastian and Maier-Hein, Lena and others},
  journal={arXiv preprint arXiv:1902.06426},
  year={2019}
}

@inproceedings{gonzalez2020isinet,
  title={ISINet: An Instance-Based Approach for Surgical Instrument Segmentation},
  author={Gonz{\'a}lez, Cristina and Bravo-S{\'a}nchez, Laura and Arbelaez, Pablo},
  booktitle={International Conference on Medical Image Computing and Computer-Assisted Intervention (MICCAI)},
  pages={595--605},
  year={2020},
  organization={Springer}
}

@inproceedings{zhao2022trasetr,
  title={TraSeTR: Track-to-Segment Transformer with Contrastive Query for Instance-Level Instrument Segmentation in Robotic Surgery},
  author={Zhao, Zixu and Jin, Yueming and Gao, Xiaojie and Dou, Qi and Heng, Pheng-Ann},
  booktitle={IEEE International Conference on Robotics and Automation (ICRA)},
  pages={11186--11193},
  year={2022},
  organization={IEEE}
}

@article{ma2024medsam,
  title={Segment Anything in Medical Images},
  author={Ma, Jun and He, Yuting and Li, Feifei and Han, Lin and You, Chenyu and Wang, Bo},
  journal={Nature Communications},
  volume={15},
  number={1},
  pages={654},
  year={2024},
  publisher={Nature Publishing Group}
}

@inproceedings{yue2024surgicalsam,
  title={SurgicalSAM: Efficient Class Promptable Surgical Instrument Segmentation},
  author={Yue, Wenxi and Zhang, Jing and Hu, Kun and Xia, Yong and Luo, Jiebo and Wang, Zhiyong},
  booktitle={Proceedings of the AAAI Conference on Artificial Intelligence},
  volume={38},
  pages={6890--6898},
  year={2024}
}

@article{paranjape2024adaptivesam,
  title={AdaptiveSAM: Towards Efficient Tuning of SAM for Surgical Scene Segmentation},
  author={Paranjape, Jay N and Nair, Nithin Gopalakrishnan and Sikder, Shameema and Vedula, S Swaroop and Patel, Vishal M},
  journal={arXiv preprint arXiv:2308.03726},
  year={2024}
}

@inproceedings{radford2021clip,
  title={Learning Transferable Visual Models From Natural Language Supervision},
  author={Radford, Alec and Kim, Jong Wook and Hallacy, Chris and Ramesh, Aditya and Goh, Gabriel and Agarwal, Sandhini and Sastry, Girish and Askell, Amanda and Mishkin, Pamela and Clark, Jack and others},
  booktitle={International Conference on Machine Learning (ICML)},
  pages={8748--8763},
  year={2021},
  organization={PMLR}
}

@inproceedings{wang2022cris,
  title={CRIS: CLIP-Driven Referring Image Segmentation},
  author={Wang, Zhaoqing and Lu, Yu and Li, Qiang and Tao, Xunqiang and Guo, Yandong and Gong, Mingming and Liu, Tongliang},
  booktitle={Proceedings of the IEEE/CVF Conference on Computer Vision and Pattern Recognition (CVPR)},
  pages={11686--11695},
  year={2022}
}

@inproceedings{luddecke2022clipseg,
  title={Image Segmentation Using Text and Image Prompts},
  author={L{\"u}ddecke, Timo and Ecker, Alexander},
  booktitle={Proceedings of the IEEE/CVF Conference on Computer Vision and Pattern Recognition (CVPR)},
  pages={7086--7096},
  year={2022}
}

@inproceedings{yang2022lavt,
  title={LAVT: Language-Aware Vision Transformer for Referring Image Segmentation},
  author={Yang, Zhao and Wang, Jiaqi and Tang, Yansong and Chen, Kai and Zhao, Hengshuang and Torr, Philip HS},
  booktitle={Proceedings of the IEEE/CVF Conference on Computer Vision and Pattern Recognition (CVPR)},
  pages={18155--18165},
  year={2022}
}

@article{marinov2024interactive,
  title={Deep Interactive Segmentation of Medical Images: A Systematic Review and Taxonomy},
  author={Marinov, Zdravko and J{\"a}ger, Paul F and Egger, Jan and Kleesiek, Jens and Stiefelhagen, Rainer},
  journal={IEEE Transactions on Pattern Analysis and Machine Intelligence},
  volume={46},
  number={12},
  pages={10998--11018},
  year={2024},
  publisher={IEEE}
}

@inproceedings{zhu2024meduhip,
  title={MedUHIP: Towards Human-In-the-Loop Medical Segmentation},
  author={Zhu, Jiayuan and Wu, Junde},
  booktitle={International Conference on Medical Image Computing and Computer-Assisted Intervention (MICCAI)},
  year={2024},
  note={arXiv preprint arXiv:2408.01620}
}

@article{bai2023qwenvl,
  title={Qwen-VL: A Versatile Vision-Language Model for Understanding, Localization, Text Reading, and Beyond},
  author={Bai, Jinze and Bai, Shuai and Yang, Shusheng and Wang, Shijie and Tan, Sinan and Wang, Peng and Lin, Junyang and Zhou, Chang and Zhou, Jingren},
  journal={arXiv preprint arXiv:2308.12966},
  year={2023}
}

@article{iglovikov2018ternausnet,
  title={Ternausnet: U-net with vgg11 encoder pre-trained on imagenet for image segmentation},
  author={Iglovikov, Vladimir and Shvets, Alexey},
  journal={arXiv preprint arXiv:1801.05746},
  year={2018}
}

@inproceedings{jin2019incorporating,
  title={Incorporating temporal prior from motion flow for instrument segmentation in minimally invasive surgery video},
  author={Jin, Yueming and Cheng, Keyun and Dou, Qi and Heng, Pheng-Ann},
  booktitle={International conference on medical image computing and computer-assisted intervention},
  pages={440--448},
  year={2019},
  organization={Springer}
}

@article{xue2022s3net,
  title={S3Net: Spectral--spatial Siamese network for few-shot hyperspectral image classification},
  author={Xue, Zhaohui and Zhou, Yiyang and Du, Peijun},
  journal={IEEE Transactions on Geoscience and Remote Sensing},
  volume={60},
  pages={1--19},
  year={2022},
  publisher={IEEE}
}

@inproceedings{ayobi2023matis,
  title={Matis: Masked-attention transformers for surgical instrument segmentation},
  author={Ayobi, Nicol{\'a}s and P{\'e}rez-Rond{\'o}n, Alejandra and Rodr{\'\i}guez, Santiago and Arbel{\'a}ez, Pablo},
  booktitle={2023 IEEE 20th International Symposium on Biomedical Imaging (ISBI)},
  pages={1--5},
  year={2023},
  organization={IEEE}
}
\bibliographystyle{icml2026}

\newpage
\appendix
\onecolumn

\end{document}